\documentclass[sigconf]{acmart}

\usepackage{adjustbox}
\usepackage{subcaption}
\usepackage{amsmath, amssymb}
\usepackage{hhline, multirow, makecell}
\usepackage{amsmath}
\usepackage{amsthm}
\AtBeginDocument{%
  \providecommand\BibTeX{{%
    \normalfont B\kern-0.5em{\scshape i\kern-0.25em b}\kern-0.8em\TeX}}}

\copyrightyear{2020}
\acmYear{2020}
\setcopyright{acmlicensed}
\acmConference[MileTS '20]{MileTS '20: 6th KDD Workshop on Mining and Learning from Time Series}{August 24th, 2020}{San Diego, California, USA}
\acmBooktitle{MileTS '20: 6th KDD Workshop on Mining and Learning from Time Series, August 24th, 2020, San Diego, Claifornia, USA}
\acmPrice{15.00}
\acmDOI{10.1145/1122445.1122456}
\acmISBN{978-1-4503-9999-9/18/06}

\begin{document}

%
\title[RSM-GAN for Anomaly Detection]{Improving Robustness on Seasonality-Heavy Multivariate Time Series Anomaly Detection}

%

\author{Farzaneh Khoshnevisan}
\affiliation{%
  \institution{North Carolina State University}
  \city{Raleigh}
  \state{NC}
}
\email{fkhoshn@ncsu.edu}

\author{Zhewen Fan}

\affiliation{%
  \institution{Intuit AI}
  \city{San Diego}
  \state{CA}
}
\email{zhewen\_fan@intuit.com}
\author{Vitor R. Carvalho}

\affiliation{%
  \institution{Intuit AI}
  \city{San Diego}
  \state{CA}
}
\email{vitor\_carvalho@intuit.com}

%

%
\begin{abstract}
Robust Anomaly Detection (AD) on time series data is a key component for monitoring many complex modern systems. These systems typically generate high-dimensional time series that can be highly noisy, seasonal, and inter-correlated. This paper explores some of the challenges in such data, and proposes a new approach that makes inroads towards increased robustness on seasonal and contaminated data, while providing a better root cause identification of anomalies. In particular, we propose the use of Robust Seasonal Multivariate Generative Adversarial Network (RSM-GAN) that extends recent advancements in GAN with the adoption of convolutional-LSTM layers and attention mechanisms to produce excellent performance on various settings. We conduct extensive experiments in which not only do this model displays more robust behavior on complex seasonality patterns, but also shows increased resistance to training data contamination. We compare it with existing classical and deep-learning AD models, and show that this architecture is associated with the lowest false positive rate and improves precision by 30\% and 16\% in real-world and synthetic data, respectively.
\end{abstract}

%
%

\ccsdesc[500]{Computing methodologies~Anomaly detection}

%
\keywords{anomaly detection, multivariate time series, seasonality-heavy data, generative adversarial networks, attention mechanism}

%
\maketitle

\section{Introduction}
\noindent Detecting anomalies in real-time data sources is critical thanks to the steady rise in the complexity of modern systems, ranging from satellite system monitoring to cyber-security. Such systems often produce multi-channel time series data that automatically detecting anomalous moments can be quite challenging to any anomaly detection (AD) system due to its intrinsic inter-correlation, seasonality, trendiness, and irregularity traits. Speedy detection, along with timely corrective measures before any catastrophic failure, are also key considerations for time-series AD systems.

 Multivariate time-series (MTS) AD on seasonality-heavy data can be challenging to most techniques proposed in the literature. Classical time-series forecasting techniques, such as Autoregressive Integrated Moving Average (ARIMA) \cite{arima} and Statistical Process Control (SPC) \cite{spc}, in general cannot adequately capture the inter-dependencies among MTS. Also, classical density or distance-based models, such as K-Nearest Neighbors (KNN) \cite{knn}, usually ignore the effect of temporal dependencies and/or seasonality in time series. In recent years, deep learning architectures have achieved great success due to their ability to learn the latent representation of normal samples, such as Auto-encoders \cite{deep-ae} and Generative Adversarial Networks (GAN) \cite{gan2}. However, such advanced AD methods suffer from high false positive rate (FPR) when applied to seasonal MTS \cite{season-fpr}. Furthermore, majority of the existing AD methods are built on an unrealistic assumption that the training data is contamination free, which is rarely the case in real-world applications.

\begin{figure}[t!]
  \centering 
  \includegraphics[width=0.83\columnwidth]{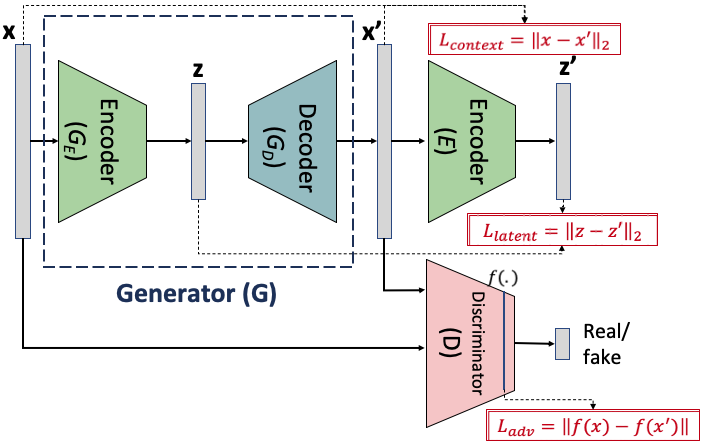}
  \caption{RSM-GAN architecture with loss definitions}
  \label{fig:GAN} 
\end{figure}

 This paper explores some of the challenges in real-world MTS, namely multi-period seasonality and training data contamination, by proposing a GAN-based architecture, named Robust Seasonal Multivariate GAN (RSM-GAN), that has an encoder-decoder-encoder structure as shown in Figure \ref{fig:GAN}. Co-training of an additional encoder enables this model to be robust against noise and contamination in training data. A novel smoothed attention mechanism is employed in recurrent layers of the encoders to account for multiple seasonality patterns in a data-driven manner. Also, we propose a causal inference framework for root cause identification. We conduct extensive empirical studies on synthetic data with various levels of seasonality and contamination, along with a real-world encryption key dataset. The results show superiority of RSM-GAN for timely and precise detection of anomalies and root causes as compared to state-of-the-art baseline models.

 Contributions of our work can be summarized as follows: (1) we propose a convolutional recurrent Wasserstein GAN architecture (RSM-GAN) that detects anomalies in MTS data precisely
; (2) we explicitly model seasonality as part of the RSM-GAN architecture through a novel smoothed attention mechanism; (3) we apply an additional encoder to handle the contaminated training data; (4) we propose a scoring and causal inference framework to accurately and timely identify anomalies and to pinpoint unspecified number of root cause(s). The RSM-GAN framework enables system operators to react to abnormalities swiftly and in real-time manner, while giving them critical information about the root cause(s) and severity of the anomalies.

\section{Related Work}
MTS anomaly detection has long been an active research area because of its critical importance in monitoring high risk tasks. 
Classical time series analysis models such as Vector Auto-regression (VAR) \cite{var}, and latent state based models such as Kalman Filters \cite{kalman} have been applied to MTS, but they are sensitive to noise and prone to misspecification. Classical machine learning methods are also widely used that can be categorized into distance-based methods such as the KNN \cite{knn}, classification-based methods such as One-Class SVM \cite{one-svm}, and ensemble methods such as Isolation Forest \cite{iforest}. These general purpose AD methods do not account for temporal dependencies nor the seasonality patterns that are ubiquitous in MTS, which lead to non-satisfactory performance in real applications. Recently, deep neural networks with architectures such as auto-encoder and GAN-based, have shown great promise for AD in various domains. Autoencoder-based models learn low-dimensional latent representations and utilize reconstruction errors as the score to detect anomalies \cite{autoencoder1,autoencoder2,autoencoder3}. GAN-based models leverage adversarial learning for mapping high-dimensional training data to the latent space and later use latent space to calculate reconstruction loss as the anomaly score \cite{ganomaly,gan3image,gan4image}.

Recurrent neural network (RNN)-based approaches have been employed for MTS AD \cite{lstmed,rnn-ad}. \cite{madgan} proposed GAN-AD, which is the first work to apply recurrent GAN-based approach to MTS anomaly detection. However, the GAN-AD architecture is not efficient for real-time anomaly detection due to costly invert mapping step while testing. Multi-Scale Convolutional Recurrent Encoder-Decoder (MSCRED) is a deep autoencoder-based AD framework applied to MTS data \cite{mscred}. MSCRED captures inter-correlation and temporal dependency between time-series by convolutional-LSTM networks and therefore, achieves state-of-the-art performance. However, non of these models account for seasonal and contaminated training data.
A few studies have addressed seasonality by applying Fourier transform, such as Seasonal ARIMA \cite{sarima}, or time-series decomposition methods \cite{fourier-season}. Such treatments are inefficient when applied to high-dimensional MTS data while they do not account for multi-period seasonality. RSM-GAN is designed to address heavy seasonality using attention mechanism, and to improve robustness to severe levels of contamination by co-training of an encoder.

\section{Methodology}\label{method}
We define an MTS as $X=(X_1,...,X_n)\in\mathbb{R}^{n\times T}$, where $n$ is the number of time series, and $T$ is the length of the training data. We aim to predict two AD outcomes: 1) the time points $t$ after $T$ that are associated with anomalies, and 2) time series $ i \in \{1,..,n\}$ causing the anomalies. 
In the following, we first describe how we transform the raw MTS to be consumed by a convolutional recurrent GAN. Then we introduce the RSM-GAN architecture and the seasonal attention mechanism. Finally, we describe anomaly scoring and causal inference procedure to identify anomalies and the root causes in the prediction phase.

\subsection{RSM-GAN Framework}
\subsubsection{MTS to Image Conversion} To extend GAN to MTS and to capture inter-correlation between multiple time series, we convert the MTS into an image-like structure through construction of the so-called multi-channel correlation matrix (MCM), inspired by \cite{song2018deep,mscred}. 
Specifically, we define multiple windows of different sizes $W=(w_1,...,w_C)$, and calculate the pairwise inner product (correlation) of time series within each window. At a specific time point $t$, we generate $C$ matrices (channels) of shape $n\times n$, where each element of matrix $S_t^c$ for a window of size $w_c$ is calculated by this formula:
\begin{equation}
    s_{ij}=\frac{\sum_{\delta=0}^{w_c}x_i^{t-\delta}\cdot x_j^{t-\delta}}{w_c}
\end{equation}
 In this work, we select windows $W=(5, 10, 30)$. This results in $3$ channels of $n\times n$ correlation matrices for time point $t$ noted as $S_t$. To convert the span of MTS into this shape, we consider a step size $p=5$. Therefore, $X$ is transformed to $S=(S_1,...,S_M)\in\mathbb{R}^{M \times n\times n\times C}$, where  $M=\lfloor\frac{T}{p}\rfloor$ steps represented by MCMs. Finally, to capture the temporal dependency between consecutive steps, we stack $h=4$ previous steps to the current step $t$ to prepare the input to the GAN-based model. Later, we extend MCM to also capture seasonality unique to MTS.

\subsubsection{RSM-GAN Architecture} 
The idea behind using a GAN to detect anomalies is intuitive. During training, a GAN utilize adversarial learning to capture the distribution of the input data. Then, if anomalies are present during prediction, the networks would fail to reconstruct the input, thus produce large losses. In most deep AD literature, the training data is explicitly assumed to be normal with no contamination. In a study, \cite{encoder} have shown that simultaneous training of an encoder with GAN improves the robustness of the model towards contamination. This is mainly because the joint encoder forces similar inputs to lie close to each other by optimizing the latent loss, and thus account for the contamination while training. To this end, we adopt an encoder-decoder-encoder structure \cite{ganomaly}, with the additional encoder, to optimize input reconstruction in both original and latent space. Specifically, in Figure \ref{fig:GAN}, the generator $G$ has autoencoder structure that the encoder ($G_E$) and decoder ($G_D$) interact with each other to minimize the contextual loss: the $l_2$ distance between input $x$ and reconstructed input $G(x)=x'$. An additional encoder $E$ is trained jointly with the generator to minimize the latent loss: the $l_2$ distance between latent vector $z$ and reconstructed latent vector $z'$. Finally, the discriminator $D$ is tasked to distinguish between the original input $x$ and the generated input $x'$. Following the recent advancements on GAN, we employ the Wasserstein GAN with gradient penalty (WGAN-GP) \cite{wgan-gp} to ensure stable gradients, avoid the collapsing mode, and thus improve the training. 
Therefore, the final objective functions for the generator and discriminator are as following:
\begin{equation}
\begin{aligned}
    L_G =  \min_{G}\min_{E} & \Big(  w_1\mathbb{E}_{x\sim p_x} \| x-x' \|_2 +  w_2 \mathbb{E}_{x\sim p_x} \| G_E(x)-E(x') \|_2 \\ & +  w_3  \mathbb{E}_{x\sim p_x}[f_\theta(x')]\Big)
\end{aligned}
\end{equation}
\begin{equation}
    L_D = \max_{\theta \in \Theta} \mathbb{E}_{x\sim p_x}[f_\theta(x)] - \mathbb{E}_{x\sim p_x} [f_\theta(x')]
\end{equation}
\noindent where $\theta$ is the discriminator's parameter and ($w_1$, $w_2$, $w_3$) are weights controlling the effect of each loss. The choice of contextual loss weight, has the largest effect on training convergence and we chose (50, 1, 1) weights for optimal training. We employ Adam optimizer to optimize the above losses for $G$ and $D$ alternatively. Each encoder in Figure \ref{fig:GAN} is composed of multiple convolutional layers, each followed by convolutional-LSTM layers to capture both spatial and temporal dependencies in input. The detailed inner structure of each component is described in Appendix \ref{sec:apx_inner}.

\subsubsection{Seasonality Adjustment via Attention Mechanism} 
\begin{figure}[t]
  \centering 
  \includegraphics[width=0.88\columnwidth]{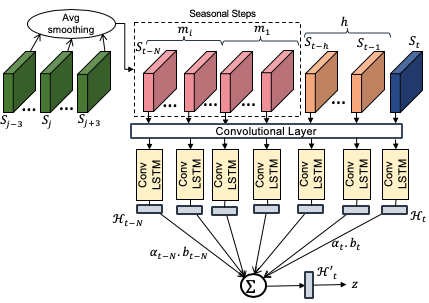} 
  \caption{Smoothed attention mechanism}
  \label{fig:attention} 
\end{figure}

In order to adjust the seasonality in MTS data, we stack previous seasonal steps to the input data, and allow the convolutional-LSTM cells in the encoder to capture temporal dependencies through an attention mechanism. Specifically, in addition to $h$ previous immediate steps, we append $m_i$ previous seasonal steps per seasonal pattern $i$. To illustrate, assume the input has both the daily and weekly seasonality. To prepare input for time step $t$, we stack MCMs of up to $m_1$ days ago at the same time, and up to $m_2$ weeks ago at the same time.  
Additionally, to account for the fact that seasonal patterns are often not exact, we smooth the seasonal steps by averaging over steps in a neighboring window of 6 steps.\\
Moreover, even though the $h$ previous steps are closer to the current time step, but the previous seasonal steps might be a better indicator to reconstruct the current step. Therefore, an attention mechanism is employed to assign weights to each step based on the similarity of the hidden state representations in the last layer using:
\begin{equation}
    \mathcal{H'}_t = \sum_{i\in (t-N,t)} \alpha_i \mathcal{H}_i  \text{, where } 
    \alpha_i=\mathrm{softmax}\Big(\frac{Vec(\mathcal{H}_t)^T Vec(\mathcal{H}_i)}{\mathcal{X}}\Big)
\end{equation}
where $N=h+\Sigma m_i$, $Vec(\cdot)$ denotes the vector, and $\mathcal{X}=5$ is the rescaling factor. Figure \ref{fig:attention} presents the structure of the described smoothed attention mechanism.
Finally, to make our model even more adaptable to real-world datasets that often exhibit holiday effects, we multiply the attention weight $\alpha_i$ by a binary bit $b_i \in \{0,1\}$, where $b_i=0$ in case of holidays or other exceptional behavior in previous steps. This way, we eliminate the effect of undesired steps from modeling the current step.

\subsection{Prediction Phase}
\subsubsection{Anomaly Score Assignment} \label{sec:AS_assign}
The residual MCM matrix from the first channel, $R_x=x_{:,:,0}-x_{:,:,0}'$, are indicative of anomalies while predicting. 
We define broken tiles as the elements of $R_x$ that have error value of greater than $\theta_b$. Previous studies have defined a scoring method based on the number of broken tiles in $R_x$ that we call context$_{b}$ \cite{mscred}. However, this score is insensitive to non-severe anomalies, and lowering the threshold would result in high FPR. Since each row/column in $R_x$ is associated with a time series, the ones with the largest number of broken tiles are contributing the most to the anomalies. Therefore, by defining a threshold $\theta_h \leq \theta_b$, we propose to only count the number of broken tiles in rows/columns with more than half broken and name this method context$_{h}$. The above thresholds $\theta = \beta \times \eta_{.996}(E_\mathrm{train})$, which is calculated based on $99.6^{th}$ percentile of error in the training residual matrices, and the best $\beta$ is chosen by a grid search on validation set.

\subsubsection{Root Cause Framework} Large errors in rows/columns of $R_x$ are indicative of anomalous behavior in those time series. To identify which are contributing the most to anomalies, we need a root-cause scoring system to assign a score to each time series based on the severity of its errors. We present two different methods: 1) number of broken tiles (NB) (using the optimized $\theta_b$), and 2) sum of absolute errors (AE). 

 Furthermore, the number of root causes, $k$, is unknown in real-world applications. Despite previous studies, we allow the elbow method \cite{elbow} to find the optimal $k$ number of time series from the root cause scores. In this approach, by sorting the scores and plotting the curve, we aim to find the point where the scores become very small and close to each other (elbow point). Time series associated with the scores greater than elbow point are identified as root causes. Thus, we define the elbow point as the point with maximum distance from a vector that connects the first and last scores.

\section{Experimental Setup}
\subsection{Data}
\textbf{Synthetic Data:} We generate synthetic sinusoidal-based multivariate time-series with different seasonal period and contamination levels to evaluate our model comprehensively. Ten time series with 2 months worth of data by minute sampling frequency are generated with length $T=80,640$. Also, anomalies with varying duration and intensity are injected to the training and test sets. The detailed data generation process in discussed in Appendix \ref{sec:apx_synth}.

\textbf{Encryption Key Data:} Our encryption-key dataset contains $7$ time series generated from a project's encryption process. Each time series represents the number of requests for a specific encryption key per minute. The dataset contains $4$ months of data with length $T=156,465$. Four anomalies with various length and scales are identified in the test sequence by a security expert, and we randomly injected $5$ additional anomalies into both the training and test sets.

\subsection{Evaluation}
RSM-GAN is compared against two classical machine learning models, i.e., One-class SVM (OC-SVM) \cite{one-svm} and Isolation Forest \cite{iforest}. and a deep autoencoder-based AD model called MSCRED \cite{mscred}. MSCRED is run in a sufficient number of epochs and its best performance is reported. We also have tried GAN-AD \cite{madgan} in multiple settings, but the results are not reported here due to its inefficient and faulty performance. For evaluation, in addition to precision, recall, FPR, and F1 score, we include the \textbf{Numenta Anomaly Benchmark (NAB)} score \cite{numenta}. NAB is a standard open source framework for evaluating real-time AD algorithms. The NAB assigns score to each true positive detection based on their relative position to the anomaly window (by a scaled \textit{sigmoid} function), and negative score to false positives and false negatives.

 In all experiments, the first half of the time series are used for training and the remainder for validation/test by 1/5 ratio. RSM-GAN is implemented in Tensorflow and trained in 300 epochs, in batches of size $32$, on an AWS Sagemaker instance with four $16$GB GPUs. 
All the results are produced by an average over five independent runs.

\section{Results and Discussion}
\subsection{Anomaly Score and Root Cause Assessment}
In this section, we first compare the performance of our RSM-GAN using the two context$_{b}$ and context$_{h}$ anomaly score assignment methods described in Section \ref{sec:AS_assign}. Table \ref{tab:scores} reports the performance on synthetic MTS with no contamination and seasonality with the optimized threshold as illustrated. As we can see, our proposed context$_{h}$ method outperforms context$_{b}$ for all metrics except of recall. Specifically, context$_{h}$ improves the precision and FPR by $6.2\%$ and $0.08\%$, respectively. Since the same result holds for other settings, we report the results based on context$_{h}$ scoring in the following experiments.

\begin{table}[tb]
\caption{Different anomaly score assignment performances}
\begin{adjustbox}{width=0.97\columnwidth,center}
\begin{tabular}{c|c|ccccc}
\textbf{Score} &\textbf{Threshold}&\textbf{Precision}& \textbf{Recall} & \textbf{F1} &\textbf{FPR} & \textbf{NAB Score} \\ \hline
context$_{b}$ & 0.0019 & 0.784 & \textbf{0.958} &	0.862 &	0.0023 &	0.813 \\
context$_{h}$ & 0.00026 &\textbf{0.846} &	0.916 &	\textbf{0.880} &	\textbf{0.0015} &	\textbf{0.859} \\
\end{tabular}
\end{adjustbox}
\label{tab:scores}
\end{table}

 Next, we compare the two root cause scoring methods for the baseline MSCRED and our RSM-GAN. Root causes are identified based on the average of $R_x$'s in an anomaly window. 
Synthetic data used in this experiment has two combined seasonal patterns and ten anomalies in training sequence. Overall, RSM-GAN outperforms MSCRED (marked by *). As the results suggest, the NB method performs the best for MSCRED. However, for RSM-GAN the AE leads to the best performance. Since the same results hold for other settings, we report NB for MSCRED and AE for RSM-GAN in subsequent sections.

\begin{table}[tb]
\caption{Different root cause identification performances}
\begin{adjustbox}{width=0.9\columnwidth,center}
\begin{tabular}{c||c|ccc}
\textbf{Model} &\textbf{Scoring}&\textbf{Precision}& \textbf{Recall} & \textbf{F1} \\ \hline
\multirowcell{2}{MSCRED} 
& Number of broken (NB) & 0.5154 &	\textbf{0.7933} &	\textbf{0.6249} \\
& Absolute error (AE) & \textbf{0.5504} &	0.7066 &	0.6188 \\ \hline

\multirowcell{2}{RSM-GAN} 
& Number of broken (NB) & 0.4960 &	0.8500 &	0.6264 \\
& Absolute error (AE) & \textbf{0.6883*} &	\textbf{0.8666*} &	\textbf{0.7672*} \\ 

\end{tabular}
\end{adjustbox}
\label{tab:rootcause}
\end{table}

\subsection{Contamination Resistance Assessment}
We assess the robustness of RSM-GAN towards different levels of contamination in training data. In this experiment, we start with no contamination and at each subsequent level, we add $5$ more random anomalies with varying duration to the training data. The percentages presented in the first column in Table \ref{tab:contamination} shows the proportions of the anomalous time points in train/test time span.  Results in Table \ref{tab:contamination} suggest that our proposed model outperforms all baseline models at all contamination levels for all metrics except of the recall. Note that the 100\% recall for classic baselines  are at the expense of FPR as high as $26.4\%$. Furthermore, comparison of the NAB score shows that our model has more timely detection and less irrelevant false positives. Lastly, as we can see, the MSCRED performance drops drastically as the contamination level increases, due to the normal training data assumption and the encoder-decoder architecture.

\begin{table*}[hbt]
\caption{Model Performance on synthetic data with different levels of training contamination and random seasonality}
\begin{adjustbox}{width=0.69\textwidth,center}
\begin{tabular}{c||c|ccccc|c}
\textbf{Contamination} &\textbf{Model}&\textbf{Precision}& \textbf{Recall} & \textbf{F1} &\textbf{FPR} & \textbf{NAB Score} & \textbf{Root Cause Recall} \\ \hline
\multirowcell{4}{No contamination \\ train: 0 (0) \\ test: 10 (\%0.82)} & OC-SVM & 0.1581 &	\textbf{1.0000} &	0.2730 &	0.0473 &	-8.4370 & - \\
& Isolation Forest & 0.0326 &	\textbf{1.0000} &	0.0631 &	0.2640 &	-51.4998 & - \\
& MSCRED & 0.8000 &	0.8450 &	0.8219 &	0.0018 &	0.7495 & \textbf{0.7533} \\
& RSM-GAN & \textbf{0.8461} &	0.9166 &	\textbf{0.8800} &	\textbf{0.0015} &	\textbf{0.8598} &	0.6333 \\ \hline

\multirowcell{4}{Mild contamination \\ train: 5 (\%0.43) \\ test: 10 (\%0.76)} 
& OC-SVM & 0.2810 &	\textbf{1.0000} &	0.4387 &	0.0218 &	-3.3411 & - \\
& Isolation Forest & 0.3134 &	\textbf{1.0000} &	0.4772 &	0.0187 &	-2.7199 & - \\
& MSCRED & 0.6949 &	0.6029 &	0.6457 &	0.0023 &	0.2721 &	0.5483 \\
& RSM-GAN & \textbf{0.8906} &	0.7500 &	\textbf{0.8143} &	\textbf{0.0009} &	\textbf{0.8865} &	\textbf{0.7700} \\ \hline

\multirowcell{4}{Medium contamination \\ train: 10 (\%0.82) \\ test: 10 (\%0.85)} 
& OC-SVM & 0.4611 &	\textbf{1.0000} &	0.6311 &	0.0113 &	-1.2351 & - \\
& Isolation Forest & 0.6311 &	\textbf{1.0000} &	0.7739 &	0.0056 &	-0.1250 & - \\
& MSCRED & 0.6548 &	0.7143 &	0.6832 &	0.0036 &	0.2712 &	0.6217 \\
& RSM-GAN & \textbf{0.8553} &	0.8442 &	\textbf{0.8497} &	\textbf{0.0014} &	\textbf{0.8511} &	\textbf{0.8083} \\ \hline

\multirowcell{4}{Severe contamination \\ train: 15 (\%1.19) \\ test: 15 (\%1.18)} 
& OC-SVM & 0.5691 &\textbf{	1.0000} &	0.7254 &	0.0102 &	-0.3365 & - \\
& Isolation Forest & 0.8425 &	\textbf{1.0000} &	\textbf{0.9145} &	0.0025 &	0.6667 & - \\
& MSCRED & 0.5493 &	0.7290 &	0.6265 &	0.0080 &	0.0202 &	0.6611 \\
& RSM-GAN & \textbf{0.8692} &	0.8774 &	0.8732 &	\textbf{0.0018} &	\textbf{0.8872} &	\textbf{0.8133} \\ \hline
\end{tabular}
\end{adjustbox}
\label{tab:contamination}
\end{table*}



\begin{table*}[htb!]
\caption{Model performance on synthetic data with different seasonal patterns and no training contamination}
\begin{adjustbox}{width=0.69\textwidth,center}
\begin{tabular}{c||c|ccccc|c}
\textbf{Seasonality} &\textbf{Model}&\textbf{Precision}& \textbf{Recall} & \textbf{F1} &\textbf{FPR} & \textbf{NAB Score} & \textbf{Root Cause Recall} \\ \hline

\multirowcell{4}{Random seasonality} 
& OC-SVM & 0.4579 &	0.9819 &	0.6245 &	0.0097 &	-8.6320 & - \\
& Isolation Forest & 0.0325 &	\textbf{1.0000} &	0.0630 &	0.2646 &	-51.606 &- \\
& MSCRED & 0.8000 &	0.8451 &	0.8219 &	0.0019 &	0.7495 & \textbf{0.7533} \\
& RSM-GAN & \textbf{0.8462} &	0.9167 &	\textbf{0.8800} &	\textbf{0.0015} &	\textbf{0.8598} & 0.6333 \\ \hline

\multirowcell{4}{Daily seasonality} 
& OC-SVM & 0.1770 &	\textbf{1.0000} &	0.3008 &	0.0532 &	-9.5465 &  - \\
& Isolation Forest & 0.1387 &	\textbf{1.0000} &	0.2436 &	0.0710 &	-13.107 & - \\
& MSCRED & 0.7347 &	0.7912 &	0.7619 &	0.0033 &	0.3775 & \textbf{0.7467} \\
& RSM-GAN & \textbf{0.9012} &	0.7935 &	\textbf{0.8439} &	\textbf{0.0010} &	\textbf{0.5175} & 0.6717 \\ \hline

\multirowcell{4}{Daily and weekly \\ seasonality} 
& OC-SVM & 0.1883 &	\textbf{0.9487} &	0.3142 &	0.0400 &	-6.9745 &  - \\
& Isolation Forest & 0.1783 &	\textbf{0.9487} &	0.3002 &	0.0428 &	-7.5278 &  - \\
& MSCRED & 0.6548 &	0.7143 &	0.6832 &	0.0036 &	0.2712 &	\textbf{0.6217} \\
& RSM-GAN & \textbf{0.9000} &	0.6750 &	\textbf{0.7714} &	\textbf{0.0008} &	\textbf{0.5461} & 0.4650 \\ \hline

\multirowcell{4}{Weekly and monthly \\ seasonality \\ with holidays} 
& OC-SVM & 0.2361 &	\textbf{0.9444} &	0.3778 &	0.0425 &	-1.7362 & - \\
& Isolation Forest & 0.2783 &	0.8889 &	0.4238 &	0.0321 &	-1.0773 &  - \\
& MSCRED & 0.0860 &	0.7059 &	0.1534 &	0.0983 &	-5.1340 & 0.6067 \\
& RSM-GAN & \textbf{0.6522} &	0.8108 &	\textbf{0.7229} &	\textbf{0.0063} &	\textbf{0.5617} & \textbf{0.8667} \\ \hline
\end{tabular}
\end{adjustbox}
\label{tab:seasonality}
\end{table*}


\begin{table*}[htb!]
\caption{Model performance on encryption key and synthetic two-period seasonal MTS with medium contamination}
\begin{adjustbox}{width=0.64\textwidth,center}
\begin{tabular}{c||c|ccccc|c}
\textbf{Dataset} &\textbf{Model}&\textbf{Precision}& \textbf{Recall} & \textbf{F1} &\textbf{FPR} & \textbf{NAB Score} & \textbf{Root Cause Recall} \\ \hline

\multirowcell{4}{Encryption \\ key} 
& OC-SVM & 0.1532 &	0.2977 &	0.2023 &	0.0063 &	-17.4715& - \\
& Isolation Forest & 0.3861 &	\textbf{0.4649} &	0.4219 &	0.0028 &	-6.9343	& - \\
& MSCRED & 0.1963 &	0.2442 &	0.2176 &	0.0055 &	-1.1047	& 0.4709 \\
& RSM-GAN & \textbf{0.6852} &	0.4405 &	\textbf{0.5362} &	\textbf{0.0011} &	\textbf{0.2992}	& \textbf{0.5093} \\ \hline

\multirowcell{4}{Synthetic} 
& OC-SVM & 0.6772 &	0.9185 &	0.7772 &	0.0038 &	-2.7621 & - \\
& Isolation Forest & 0.7293 &	\textbf{0.9610} &	0.8221 &	0.0033 &	-2.2490 & - \\
& MSCRED & 0.6228 &	0.7403 &	0.6746 &	0.0043 &	0.2753 &	0.6600 \\
& RSM-GAN & \textbf{0.8884} &	0.8438 &	\textbf{0.8649} &	\textbf{0.0010} &	\textbf{0.8986} &	\textbf{0.7870} \\ \hline

\end{tabular}
\end{adjustbox}
\label{tab:realworld}
\end{table*}

\subsection{Seasonality Adjustment Assessment}
 Next, we assess the performance of our proposed attention mechanism, assuming no training contamination exists. In the first experiment, synthetic MTS includes 2 months of data, sampled per minute, with only random seasonality. Daily and weekly seasonality patterns are added at each further step. In the last experiment, we simulate $3$ years of hourly data, and add special patterns to illustrate holiday effect in both training and test data. The test set of each experiment is contaminated with $10$ random anomalies. Comparing the results in Table \ref{tab:seasonality}, RSM-GAN shows consistent performance due to the attention adjustment strategy. All the other baseline models, especially MSCRED's performance deteriorate with increased complexity of seasonal patterns.
In the last experiment in Table \ref{tab:seasonality}, all of the abnormalities injected to holidays are misidentified by the baseline models as anomalies, since no holiday adjustment is incorporated in those models and thus, low precision and high FPR has emerged. In RSM-GAN, multiplying the binary vectors of holidays with the attention weights enables accountability for extreme events and leads to the best performance in almost all metrics. 

\subsection{Performance on Real-world Dataset}
\noindent This section evaluates our model on a real-world encryption key dataset that has both daily and weekly seasonality. To be comprehensive, we also create a synthetic dataset with similar seasonality patterns and 10 anomalies in the training set. 
From Table \ref{tab:realworld}, we make the following observations: 1) RSM-GAN consistently outperforms all the baseline models for anomaly detection and root cause identification recall in both datasets. 2) Not surprisingly, for all the models, performance on the synthetic data is better than that of encryption key data. It is due to the excessive irregularities and noise in the encryption key data. 
\begin{figure}[bt]
  \centering 
  \includegraphics[width=\columnwidth]{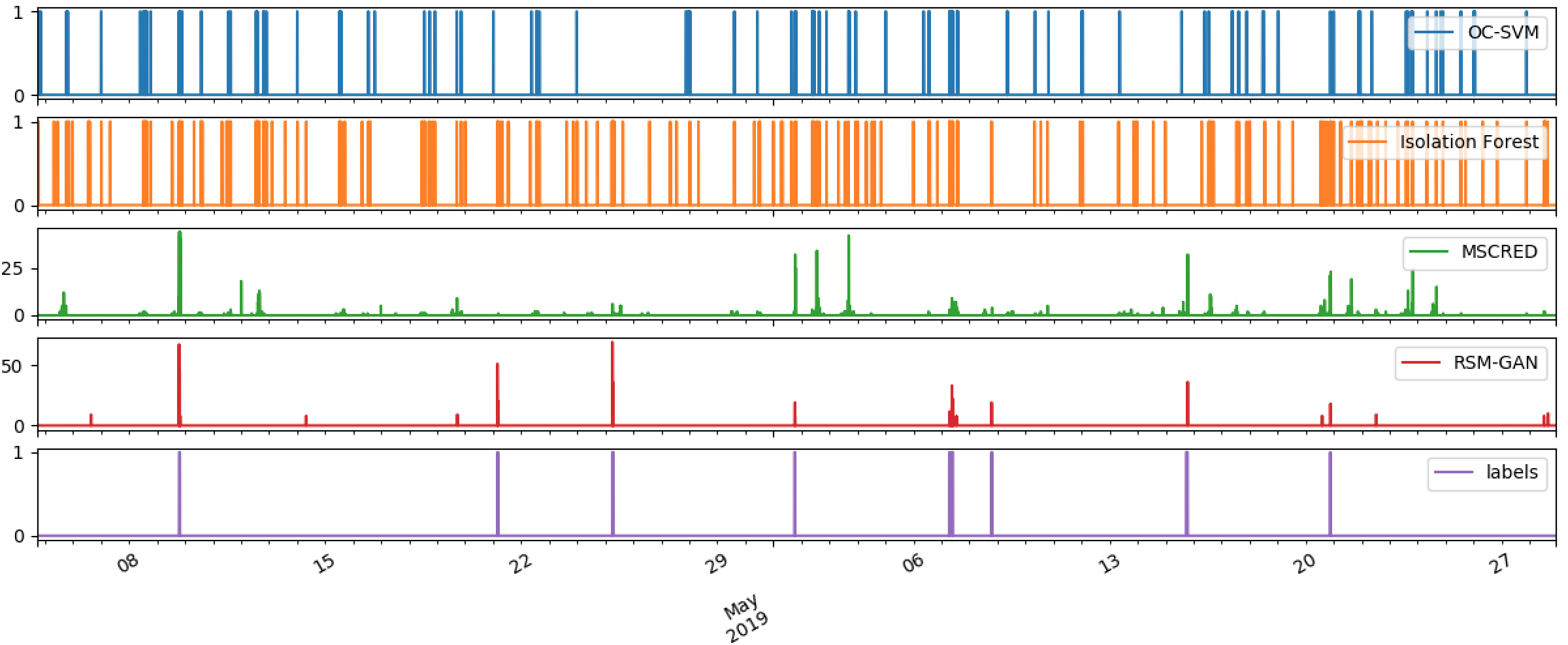}
   \caption{Anomaly score assignment on encryption key data}
   \label{fig:final_plot}
\end{figure}
3) Figure \ref{fig:final_plot} illustrates the anomaly scores assigned to each time point in test dataset by each algorithm, with the bottom plot presenting the ground truth. As we can see, even though isolation forest has the highest recall rate, it also detects many false positives not related to the actual anomaly windows, leading to negative NAB scores. 4) By comparing our model to MSCRED in Figure \ref{fig:final_plot}, we can see that MSCRED not only has much higher FPR, but it also fails to capture some anomalies. 

\section{Conclusion}
In this work, we proposed a GAN-based AD framework to handle contaminated and seasonality-heavy multivariate time-series. RSM-GAN leverages adversarial learning to accurately capture temporal and spatial dependencies in the data, while simultaneously training an additional encoder to handle training data contamination. The novel attention mechanism in the recurrent layers of RSM-GAN enables the model to adjust complex seasonal patterns often found in the real-world data. We conducted extensive empirical studies and results show that our architecture together with a new score assignment and causal inference lead to an exceptional performance over advanced baseline models on both synthetic and real-world datasets.

%
\bibliographystyle{ACM-Reference-Format}
\bibliography{kdd20}

\appendix
\section{Inner Structure of Encoder and Decoder}\label{sec:apx_inner}
\begin{figure}[tb]
  \centering 
  \includegraphics[width=0.52\textwidth]{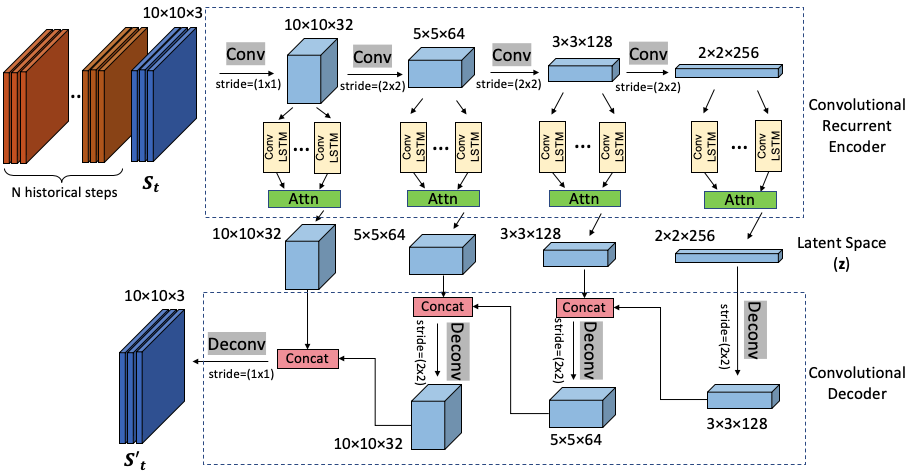} 
  \caption{Inner structure of convolutional recurrent encoders and convolutional decoder (with $n=10$)}
  \label{fig:ED_internal} 
\end{figure}

Figure \ref{fig:ED_internal} illustrates the detailed inner-structure of the encoder and decoder networks. The convolutional recurrent encoder takes the stack MCMs as input and process them by four layers of convolutional networks. Aditionally, we add RNN layers to jointly capture the spatial and temporal patterns of our MCM inputs by using convolutional-LSTM (convLSTM) cells. We apply convLSTM to the output of every convolutional layer due to its optimal mapping to the latent space. An attention mechanism is implemented inside the hidden unit of each convLSTM network. The convolutional decoder applies multiple deconvolutional layers in reverse order to reconstruct the MCM related to $x$. Starting from the last convLSTM output, decoder applies deconvolutional layer and concatenates the output with convLSTM output of the previous step from the encoder. The concatenation output is further an input to the next deconvolutional layer, and so on as illustrated in Figure \ref{fig:ED_internal} (for more detailed description of the encoder-decoder structure, refer to \cite{mscred}). The second encoder $E$ follows the same structure as the generator's encoder $G_E$ to reconstruct latent space $z'$. Input to the discriminator is the original and the generated MCM ($x$ and $x'$) at each time step. Therefore, the discriminator is a convolutional neural network with three layers, while the last layer represents $f(\cdot)$.

\section{Synthetic Data Generation}\label{sec:apx_synth}
To simulate data with different seasonality and contamination, we first generate sinusoidal-based waves of length $T$ and periodicity $F$:
\begin{equation}
    S(t, F) = \left\{
    \begin{array}{ll}
        \sin[(t-t_0)/F]+0.3\times\epsilon_t & s_{rand}=0 \\
        \cos[(t-t_0)/F]+0.3\times\epsilon_t & s_{rand}=1
    \end{array}
\right.
\end{equation}
Where $s_{rand}$ is 0 or 1, $t_0 \in [10,100]$ is shift in phase and they are randomly selected for each time series. $\epsilon_t \sim \mathcal{N}(0,1)$ is the random noise. Ten time series with 2 months worth of data by minute sampling frequency are generated, or $T=80,640$. Each time series with combined seasonality is generated by:
\begin{equation}
    S(t) = S(t, F_{rand})+ S(t, F_{day}) + S(t,F_{week})
\end{equation}
Where $F_{day} = \frac{2\pi}{60\times24}$, $F_{week} = \frac{2\pi}{60\times24\times7}$, and $F_{rand} \in [60,100]$. To simulate anomalies with varying duration and intensity, we shock time series with a random duration (within $[5,60]$ minutes), direction (spikes or dips), and number of root causes (within $[2,6]$). 
We conduct each experiment with different seasonality pattern and contamination setting.\\

\end{document}